\documentclass{article} 
\usepackage{iclr2027_conference,times}

\usepackage{amsmath,amsfonts,bm}

\def\eqref#1{equation~\ref{#1}}

\def\1{\bm{1}}

\def\rvepsilon{{\mathbf{\epsilon}}}

\def\vb{{\bm{b}}}

\def\vf{{\bm{f}}}

\def\vm{{\bm{m}}}

\def\vu{{\bm{u}}}
\def\vv{{\bm{v}}}
\def\vw{{\bm{w}}}
\def\vx{{\bm{x}}}

\def\vz{{\bm{z}}}

\def\mI{{\bm{I}}}

\DeclareMathAlphabet{\mathsfit}{\encodingdefault}{\sfdefault}{m}{sl}
\SetMathAlphabet{\mathsfit}{bold}{\encodingdefault}{\sfdefault}{bx}{n}

\usepackage[hidelinks]{hyperref}
\usepackage{url}
\usepackage{graphicx}
\usepackage[utf8]{inputenc} 
\usepackage{booktabs}       
\usepackage{amsfonts}       
\usepackage{nicefrac}       
\usepackage{microtype}      
\usepackage{amsmath}
\usepackage{subcaption}  
\usepackage[ruled,vlined,linesnumbered]{algorithm2e}
\usepackage{multirow}
\usepackage{siunitx}
\usepackage{float}
\usepackage[table]{xcolor}
\usepackage{xspace}
\usepackage{wrapfig}
\usepackage{listings}
\usepackage{enumitem}
\definecolor{codeblue}{rgb}{0.25,0.5,0.5}
\definecolor{codekw}{rgb}{0.85, 0.18, 0.50}
\definecolor{codesign}{RGB}{0, 0, 255}
\definecolor{codefunc}{rgb}{0.85, 0.18, 0.50}
\definecolor{coflowdefault}{RGB}{230,242,248}
\usepackage{caption}

\lstdefinelanguage{PythonFuncColor}{
  language=Python,
  alsoletter={_},
  emph={rand_t, randn_like, randperm, mse_loss, fn}, 
  emphstyle=\color{codefunc}, 
  keywordstyle=\color{blue}\bfseries,
  commentstyle=\color{codeblue}\itshape, 
  stringstyle=\color{orange},
  showstringspaces=false,
  basicstyle=\ttfamily\small,
  literate=
    {*}{{\color{codesign}* }}{1}
    {-}{{\color{codesign}- }}{1}
    {+}{{\color{codesign}+ }}{1}
}

\usepackage{tabularx}

\definecolor{ablSignal}{RGB}{252,230,230}
\definecolor{ablPolicy}{RGB}{255,237,179}
\definecolor{ablStrength}{RGB}{229,240,250}

\definecolor{ablSignalTitle}{RGB}{252,230,230}
\definecolor{ablPolicyTitle}{RGB}{255,237,179}
\definecolor{ablStrengthTitle}{RGB}{229,240,250}

\newcommand{\ablationheading}[2]{%
    \begingroup
    \setlength{\fboxsep}{1.5pt}%
    \colorbox{#1}{\strut\textbf{#2}}%
    \endgroup
}
\usepackage{color}
\definecolor{citecolor}{RGB}{66,168,235}
\definecolor{linkcolor}{RGB}{255,0,0}
\hypersetup{colorlinks=true,citecolor=citecolor,linkcolor=linkcolor}

\title{Smoother Flow Matching via Contrastive Trajectory Repulsion}

\author{Ziqi Jiang, Zhenqi He, and Long Chen$^\dagger$ \\
Department of CSE, The Hong Kong University of Science and Technology\\
\texttt{zjiangbl@connect.ust.hk, zheci@connect.ust.hk, longchen@ust.hk} \\
}

\newcommand{\ie}{\textit{i.e.}}
\newcommand{\eg}{\textit{e.g.}}

\iclrpreprintcopy 

\begin{document}

\maketitle

\begin{figure}[htbp] 
    \centering

    \includegraphics[width=1 \linewidth]{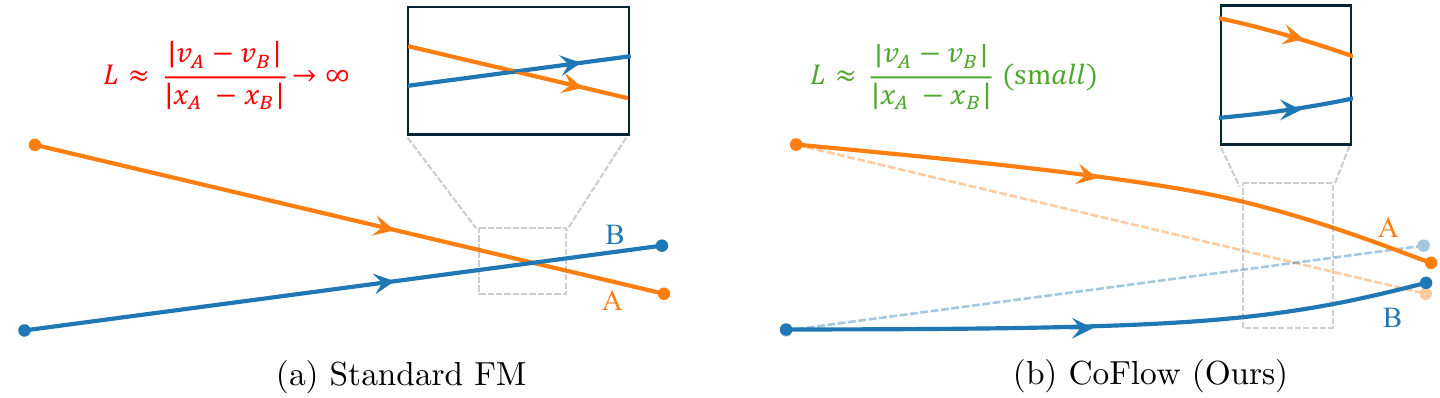}
    
    \caption{ \textbf{Illustration of our motivation.} (a) In standard FM, straight trajectories inevitably cross. At the intersection, spatially adjacent points have conflicting velocity targets ($v_A \neq v_B$ while $x_A \approx x_B$), leading to an extremely high local Lipschitz constant $L$. (b) CoFlow explicitly repels trajectories to alleviate crossing, keeping $x_A$ and $x_B$ separated. This significantly reduces the local Lipschitz constant, yielding a smoother target velocity field that is easier to learn and induces lower numerical integration errors during few-step sampling.}
    \label{fig:1}
\end{figure}

\begin{abstract}
    Trajectory crossing remains a critical bottleneck in Flow Matching (FM), and previous works typically view these crossings from a theoretical optimization perspective causing velocity averaging.  They attempt to address it indirectly by post-hoc distillation or endpoint coupling, without explicitly regulating the intermediate trajectories. In this paper, we introduce a new network learning perspective: crossing points inherently induce large local Lipschitz constants in the target velocity field, leading to two drawbacks. First, high Lipschitz constants correspond to high-frequency signals in the velocity field that neural networks struggle to fit due to spectral bias. Second, they also imply drastic velocity variations, leading to severe numerical integration errors in few-step inference.  To alleviate this,  we propose CoFlow, a framework that introduces the contrastive learning paradigm into FM to explicitly repel trajectories during training, thereby lowering the local Lipschitz constants of the velocity field. Specifically, we formulate CoFlow from a Stochastic Differential Equation (SDE) perspective by injecting a repulsive drift term. This drift actively guides the forward process of ``positive samples'' away from ``negative trajectories'', effectively reducing the local Lipschitz constant. Furthermore, we derive an equivalent stochastic interpolant formulation from this SDE, providing a simple and tractable design space to control the influence of negative samples. Extensive experiments on ImageNet 256$\times$256 demonstrate that CoFlow significantly reduces FID compared to standard FM in few-step inference (e.g., 20 steps), with no added training overhead. The code can be accessed from: 
    \url{https://github.com/HKUST-LongGroup/CoFlow}
\end{abstract}

\section{Introduction}
Continuous-time generative models, formulated as stochastic differential equations (SDEs) or ordinary differential equations (ODEs), have driven remarkable breakthroughs in high-quality visual synthesis. Specifically, diffusion models~\citep{ho2020denoising, song2020score} first gradually corrupt data with noise according to a predefined forward SDE and then learn a score function to reverse this process. However, probability paths induced by these forward SDEs are highly curved, thus requiring hundreds of sampling steps during inference. To overcome this, recent flow matching (FM)~\citep{lipman2022flow,liu2022flow, albergo2022building}  enables the construction of straight probability paths and directly learns a velocity field, substantially reducing the required sampling steps.

Although designed to straighten paths for rapid sampling, standard flow matching still requires numerous steps (e.g., 50--100) in practice to generate high-fidelity samples. From a theoretical optimization perspective, this gap stems from trajectory crossings~\citep{liu2022flow}: when trajectories cross\footnote{While exact intersections are rare in high-dimensional spaces, we use ``crossing'' in this paper to describe scenarios where distinct trajectories are close to each other but have divergent target velocities.}, the marginal velocity at intersections inherently becomes the expectation of conflicting velocities, yielding abrupt changes in the velocity field. During inference, these drastic velocity variations will amplify the truncation errors of numerical ODE solvers, thus degrading few-step generation performance. To alleviate this issue,  prevailing approaches primarily operate on the boundaries: they either rely on post-hoc distillation to bypass intermediate integration~\citep{yin2024one,song2023consistency,salimans2022progressive, liu2022flow}, or optimize endpoint couplings to reduce crossings~\citep{tong2023improving,davtyan2026faster}.  While both approaches are effective,  a more straightforward direction seems to remain unexplored: \emph{why not fix the simple couplings and directly adjust the intermediate trajectories during training to explicitly mitigate crossings?}

To bridge this gap, we first examine trajectory crossing from a neural network learning perspective, and use the local Lipschitz constant to quantify the \textbf{smoothness} of the velocity field. As illustrated in Fig.~\ref{fig:1}(a), when trajectories cross, spatially adjacent points have drastically different velocity targets, causing the Lipschitz constant of the target velocity field to be extremely large. This large Lipschitz constant leads to two critical drawbacks: 1) It corresponds to high-frequency signals in the velocity field, which neural networks struggle to fit due to spectral bias~\citep{rahaman2019spectral}. This will make the velocity field difficult to learn accurately. 2) Theoretically, a high Lipschitz constant directly amplifies the truncation error of ODE solvers,  degrading few-step sampling quality.

\begin{wrapfigure}{r}{0.3\textwidth}
  \vspace{-1.0em}
  \centering
  \includegraphics[width=\linewidth]{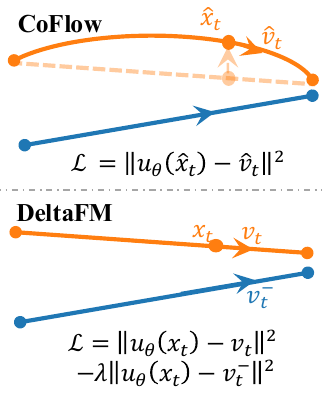}
  \caption{Comparison between DeltaFM and CoFlow.}
  \label{fig:deltafm_vs_coflow}
  \vspace{-1.0 em}
  \label{fig:deltafm}
\end{wrapfigure}

\begin{figure}[t] 
    \centering
    \includegraphics[width=\linewidth]{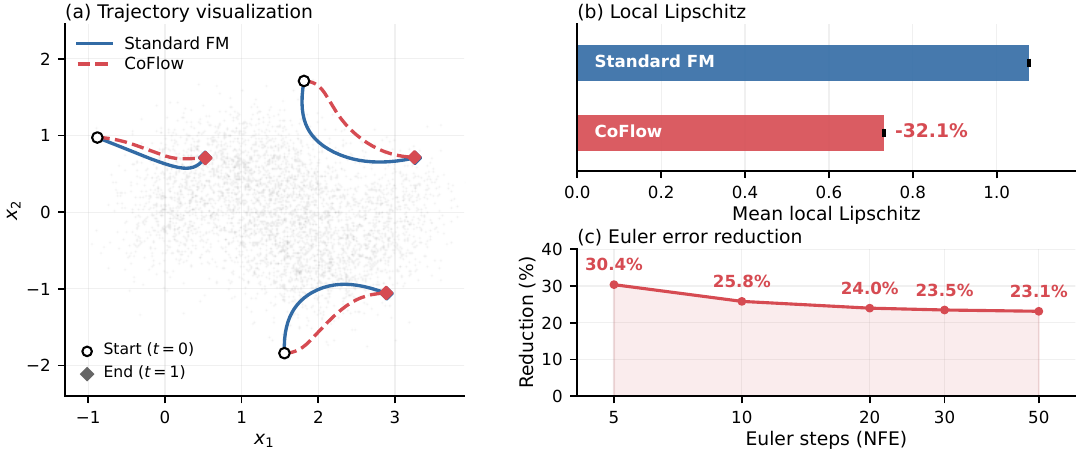}
    \vspace{-1em}
    \caption{ \textbf{2D Example: Gaussian to Two Moons.} 
(a) Visualization of generated trajectories: CoFlow yields paths with reduced curvature compared to standard FM. 
(b) Local Lipschitz constant of the learned velocity field. (c) Relative Truncation Error reduced compared with standard FM. }
    \label{fig:2d}
    \vspace{-1em}
\end{figure}

Building upon this perspective,  a compelling direction emerges: Explicitly repelling these conflicting trajectories during training to suppress the Lipschitz constant should enable the network to learn a smoother (\ie, with a lower local Lipschitz constant, see Fig.~\ref{fig:1}(b)) velocity field. Motivated by this, we introduce \textbf{CoFlow}, a novel framework that systematically rectifies training trajectories via negative samples. Drawing inspiration from contrastive learning~\citep{chen2020simple,he2020momentum}, CoFlow explicitly forces the trajectory of a target sample (positive) to repel adjacent negative paths (\eg, paths targeting other data points in a mini-batch). Theoretically, we demonstrate that this dynamic is mathematically equivalent to injecting a repulsive drift into the forward stochastic process, closely related to Doob's $h$-transform~\citep{zhou2023denoising}. From this SDE we can further derive a closed-form, stochastic interpolant (SI) formulation~\citep{albergo2022building}, which provides a flexible design space to control the influence of negative samples. We further prove that, when the repulsive drift vanishes at both endpoints, CoFlow still maps the source distribution to the target distribution, as in standard flow matching. A 2D experiment (Fig.~\ref{fig:2d}) provides some initial evidence for our motivation, showing that CoFlow can learn a velocity field with lower Lipschitz and decrease the truncation errors during infernce.


A closely related contemporary work is DeltaFM~\citep{stoica2025contrastive}, which also explores negative samples by introducing an auxiliary contrastive loss to the standard objective to penalize velocity similarity with negative samples. Despite sharing a similar motivation, DeltaFM focuses on velocity-level alignment rather than altering intermediate trajectories, as shown in Fig.~\ref{fig:deltafm}. Notably, another theoretical contribution of our work is proving that the optimal velocity field learned by DeltaFM is mathematically equivalent to an inference-time affine transformation  of the standard FM velocity field. Leveraging this insight, we can seamlessly incorporate this transformation into CoFlow at inference time, achieving  further performance gains without extra training overhead.

To further validate the effectiveness of CoFlow, we conduct extensive experiments on the ImageNet-256 generation benchmark~\citep{deng2009imagenet}. Our evaluations demonstrate that CoFlow significantly enhances the fidelity of few-step(e.g., 20 steps) image generation. Furthermore, we illustrate its compatibility with complementary techniques, such as REPA~\citep{yu2024representation}, OT-CFM~\citep{tong2023improving}, and DeltaFM~\citep{stoica2025contrastive}. In summary, our main contributions are as follows:
\vspace{-0.8em}
\begin{enumerate}[label=\textbf{\arabic*)}, leftmargin=*]
\itemsep-0.2em

\item \textbf{A Lipschitz perspective on trajectory crossing.} We quantify trajectory crossing via the local Lipschitz constant, showing that its explosion amplifies numerical truncation errors and thereby degrades few-step sampling quality.

\item \textbf{CoFlow framework inspired by Doob-\textit{h} transform.} We propose CoFlow, which explicitly repels conflicting trajectories via negative samples during training to suppress local Lipschitz constants, thus learning a smoother velocity field. We derive an exact closed-form formulation that enables efficient training without costly numerical simulation.

\item \textbf{Theoretical insight into DeltaFM.} We theoretically prove that DeltaFM is mathematically equivalent to an inference-time affine transformation, revealing its machanism and allowing it to be combined with CoFlow for further performance gains.
\end{enumerate}

\section{Related Works}


\noindent\textbf{Few Step Generation.} Improving generation quality  with few sampling steps is an important goal in diffusion and flow-based models. Existing efforts to achieve this can be broadly categorized into three paradigms. The first paradigm focuses on optimizing noise-data coupling during training to explicitly reduce trajectory crossing, such as Minibatch OT-CFM~\citep{tong2023improving}. The second paradigm relies on post-hoc rectification or distillation. For example, ReFlow~\citep{liu2022flow} and InstaFlow~\citep{liu2023instaflow} generate new synthetic data-noise pairs to retrain the model, while distillation methods like DMD~\citep{yin2024one,salimans2022progressive,song2023consistency} attempt to directly distill a pre-trained model for one-step generation.  Recently, a third paradigm has emerged, which introduces the concept of mean velocity by incorporating extra conditioning to naturally enable single-step sampling, like Shortcut Model~\citep{frans2024one} or MeanFlow~\citep{geng2025mean}.CoFlow explores a complementary approach: directly adjusting intermediate training trajectories using negative samples to encourage a smoother velocity field with lower local Lipschitz constants, thereby supporting more accurate few-step sampling.

\noindent\textbf{Contrastive Learning in Generative Models.} Contrastive Learning~\citep{he2020momentum,chen2020simple} is a powerful learning paradigm in deep learning, fundamentally relying on negative samples to benefit training. While extensively utilized in models like GANs~\citep{goodfellow2014generative}, its usage into the training of continuous-time generative models has remained relatively underexplored. Interestingly, at inference time, Classifier-Free Guidance (CFG)~\citep{ho2022classifier,dhariwal2021diffusion} acts as a \textit{de facto} contrastive mechanism, pushing the target conditional generation trajectories away from unconditional ones. The remarkable effectiveness of CFG strongly implies that explicitly incorporating contrastive learning into the training phase holds great potential. Pioneering work includes Drifting Models~\citep{deng2026generative} and DeltaFM~\citep{stoica2025contrastive}.

\section{Method}



In this section, we detail the formulation of CoFlow, a framework that incorporates negative samples into continuous-time generative modeling. Typically, continuous-time generative models can be described from two complementary perspectives: the stochastic diffusion perspective (e.g., score SDEs~\citep{song2020score}) and the stochastic interpolant(SI) perspective~\citep{albergo2022building}, which directly constructs the conditional probability path. Currently, SI has emerged as the prevailing paradigm~\citep{ma2024sit}, as it provides a more flexible design space. However, it often obscures the underlying physical intuition behind the stochastic diffusion dynamics. 

Therefore, to rigorously introduce our method, we leverage the connection between these two paradigms. Specifically, we first introduce the mechanism of CoFlow from the SDE perspective, providing an intuitive motivation for injecting a repulsive drift via negative samples. Subsequently, we derive its equivalent formulation within the stochastic interpolant framework. Finally, we demonstrate how to train a velocity field in our framework.

\subsection{Preliminaries}
\label{3.1}
\noindent\textbf{Continuous-Time Generative Modeling with SDE.} From the stochastic dynamics perspective, continuous-time generative models  define a forward process that gradually corrupts the data distribution $p_0$ to a prior noise distribution $p_1$ over time $t \in [0,1]$, governed by the following SDE:
\begin{equation}
    \label{eq1}
    \mathrm{d}\vx_t = \vf(\vx_t,t) \, \mathrm{d}t + g(t) \, \mathrm{d}\vw_t,
\end{equation}
where $\vx_t \in  \mathbb{R}^d$ represents the state at time $t$, $\vw_t$ denotes a standard $d$-dimensional Wiener process, $\vf(\cdot, t) : \mathbb{R}^d \to \mathbb{R}^d$ and $g(t) : [0,1] \to \mathbb{R}$ are the drift and diffusion coefficients, respectively.

In mainstream diffusion frameworks~\citep{ho2020denoising,song2019generative}, the drift term  is designed to be linear with respect to $\vx_t$, i.e., $\vf(x_t,t) = \alpha(t)\vx_t$. This linearity guarantees that the intermediate state $\vx_t$ can be sampled in closed form by solving Eq.~(\ref{eq1}) :
\begin{align}
    \label{eq2}
    \vx_t &= s(t)\vx_0 +  s(t)\sigma(t)\vx_1, 
\end{align}
where $\vx_0 \sim p_0$,  $ \vx_1 \sim p_1 = \mathcal{N}(0,\mI)$, and $s(t)$, $\sigma(t)$ are related schedule functions in EDM~\citep{karras2022elucidating}(More details in Appendix~\ref{appendixA}). This makes it feasible to sample $\vx_t$ in the training of the score function $\nabla_{\vx_t } \log p_t(\vx_t) $, with the ground truth as $\frac{-\vx_1}{s(t)\sigma(t)}$.  Once a score function network is learned, generation is performed by solving the corresponding reverse-time SDE from $t = 1$ to $ t = 0$. 
\begin{equation}
     \mathrm{d}\vx_t = \left [\vf(\vx_t,t) - g^2(t)\nabla_{\vx_t}\log p_t(\vx_t) \right ]\, \mathrm{d}t + g(t) \, \mathrm{d}\bar{\vw}_t,
\end{equation}
where $\mathrm{d}\bar{\vw}_t$ is a  Wiener process running backwards in time. Because directly constructing the conditional probability paths (as in Eq.~(\ref{eq2})) offers greater 
engineering efficiency, subsequent frameworks like FM opt to define the generative process directly via stochastic interpolant~\citep{albergo2023stochastic}.

\noindent \textbf{Flow Matching Configuration.} In this paper, we consider a specific stochastic-interpolant configuration where the drift and diffusion coefficients are given by $\vf(\vx_t, t) = \frac{-1}{1-t}\vx_t$ and the diffusion coefficient is $g(t) = \sqrt{\frac{2t}{1-t}}$. Under this specific SDE, the stochastic interpolant formulation of $\vx_t$ essentially becomes a simple linear interpolation between the data point and the prior noise: $\vx_t = (1-t)\vx_0 + t \vx_1$, where $s(t) = 1-t$, $\sigma(t) = \frac{t}{1-t}$.  This setup essentially corresponds to standard Flow Matching.

\noindent \textbf{Lipschitz Constant and Truncation Error.} 
The spatial smoothness of a velocity field $\vv(\vx, t)$ is characterized by its Lipschitz constant $L$, where $\|\vv(\vx_A, t) - \vv(\vx_B, t)\| \le L \|\vx_A - \vx_B\|$. When simulating the ODE via an Euler step with step size $\Delta t$, the single-step truncation error is defined as the gap between the true trajectory and the numerical update:
\begin{equation}
\mathcal{E}(\vx_t, \Delta t, \vv_t) = \|\vx(t + \Delta t) - (\vx(t) + \Delta t \, \vv(\vx(t), t))\| \le (LM+B)\Delta t^2
\end{equation}
 Where $M, B$ are constants bounding the velocity and time variable (See Appendix~\ref{app:euler_error} for detailed proof). Consequently, for a fixed step size $\Delta t$, suppressing the local Lipschitz constant $L$ effectively decreases this error, thus improving few-step sampling performance. 

\subsection{Fixed Endpoints via Doob's \textit{h}-Transform} 

The standard forward diffusion trajectory described in Eq.~(\ref{eq1}) transports a source point $\vx_0$ towards a random destination in the prior $p_1$. However,  in many practical scenarios, such as image-to-image translation~\citep{zhou2023denoising,chung2022diffusion,liu20232,kawar2022denoising},  the source-target pairs are explicitly defined. Thus, it becomes necessary to modify the forward trajectory such that it deterministically terminates at a specific target state $\vx_1$. 

This can be achieved via Doob's $h$-transform~\citep{zhou2023denoising}. Specifically, conditioning the SDE on a fixed terminal endpoint $\vx_1 $ at $t = 1$ introduces an additional drift term to Eq.~(\ref{eq1}), yielding the conditional bridge SDE that terminates at $\vx_1$:
\begin{equation}
       \mathrm{d}\vx_t = \left [\vf(\vx_t, t) + g^2(t)\cdot \nabla_{\vx_t} \log p(\vx_1|\vx_t) \right ]\, \mathrm{d}t + g(t) \, \mathrm{d}\vw_t.
\end{equation}
In this formulation, the additional drift, $g^2(t) \nabla_{\vx_t} \log p ({\vx}_1 | {\vx}_t)$,  acts as an \textbf{attractive force} guiding the particle's trajectory towards the target ${\vx}_1$.  For instance, under the specific linear configuration discussed in Sec.~\ref{3.1}, this attractive force simplifies to $ \frac{\vx_1 -\vx_t}{1-t}$. As $t \rightarrow 1$, the magnitude of this force increases to infinity, mathematically guaranteeing that the trajectory terminates exactly at $\vx_1$.

This mechanism inspires our core motivation: \textit{if injecting a specific drift can exert an attractive force towards a target point, it is a natural extension to introduce a repulsive force that pushes the trajectory away from a negative sample.}  This physical intuition forms the foundational motivation for our following CoFlow framework.

\subsection{CoFlow}
\label{3.3}
 Consider a target trajectory that transports a source point $\vx_0$ towards a destination $\vx_1$ alongside a reference pair $(\mathbf{\vx}_0^r, \mathbf{\vx}_1^r)$. We regard the reference pair as a negative sample when its induced trajectory lies in close spatial proximity to the positive trajectory, as such cases can lead to a more challenging optimization landscape, characterized by a large local Lipschitz constant. The core motivation of CoFlow is to inject a repulsive force $\vb(\vx_t,t, \vx^r_t ) $ into the drift term of Eq.~(\ref{eq1}) to push the target particle away from the negative trajectory. Formally, this repulsive SDE is defined as follows:
\begin{equation}
\label{eq5}
     \mathrm{d}\hat{\vx}_t = \left [ \vf(\vx_t,t) - \vb(\vx_t,t, \vx^r_t ) \right ]\, \mathrm{d}t + g(t) \, \mathrm{d}\vw_t
\end{equation}
Mathematically, the corresponding stochastic interpolant for this modified SDE can be expressed in the following equation (See Appendix~\ref{AppendixB} for more details):
\begin{equation}
\label{eq6}
    \hat{\vx}_t = s(t)\vx_0 + s(t)\sigma(t)\vx_1 - s(t)\int_0^t\frac{\vb(\vx_\xi,\xi,\vx^r_\xi)}{s(\xi)}\mathrm{d}\xi,
\end{equation}
where $\hat{\vx}_t$ represents the new trajectory that dynamically bypasses the negative sample. \textit{Eq.~(\ref{eq6}) demonstrates that injecting a repulsive drift into the SDE is fundamentally equivalent to modifying the standard stochastic interpolant.} 

\noindent \textbf{Simulation-Free Reparameterization.} The integral in Eq.~(\ref{eq6}) renders the sampling of $\hat{\vx}_t$ computationally intractable during training. To circumvent this, we propose a simple reparameterization trick to make it simulation-free. Specifically, instead of defining the repulsive drift $\vb$ and computing the integral, we directly define a tractable modified trajectory $\hat{\vx}_t$:
\begin{equation}
\label{eq7}
      \hat{\vx}_t = s(t)\vx_0 + s(t)\sigma(t)\vx_1 - \hat{\vb}(\vx_t,t,\vx^r_t),
\end{equation}
Here, $\vx_t = s(t)\vx_0 + s(t)\sigma(t)\vx_1$ denotes the standard baseline trajectory. Under this formulation, the repulsive drift in Eq.~(\ref{eq5}) can be analytically calculated through
\begin{equation}
\vb(\vx_t, t, \vx_t^r) = s(t) \frac{\mathrm{d}}{\mathrm{d}t} \big[ \hat{\vb}(\vx_t, t, \vx_t^r) / s(t) \big] 
\end{equation}
Crucially, CoFlow provides the flexibility to explicitly design $\hat{\vb}$. For example, by constraining $\hat{\vb}$ to vanish at the boundaries (i.e., $\hat{\vb}(\cdot, 0, \cdot) = \hat{\vb}(\cdot, 1, \cdot) = 0$), CoFlow mathematically guarantees a strictly unbiased transformation between the prior noise and the target data distribution (See proof in Appendix~\ref{AppendixC}). In this paper, we primarily focus on this unbiased regime to ensure a fair comparison with standard FM. However, the empirical success of CFG indicates that introducing specific distributional biases can sometimes yield superior generative fidelity. This suggests a broader design space for $\hat{\vb}$, potentially allowing for intentional shifts at the boundary distributions. And we leave this for future work. After deciding the drift $\hat{\vb}$, a velocity field $\vu^\theta_t(\hat{\vx}_t)$ can be learned by regressing against the closed-form target conditional velocity field: 
\begin{equation}
    \mathcal{L}_{CoFlow} = \mathbb{E} \left[ || \vu^\theta_t(\hat{\vx}_t) - \vv_t(\hat{\vx}_t) ||^2 \right] 
\end{equation}
where the input $\hat{x}_t$ is sampled from Eq.~(\ref{eq7}),  the target $v_t(\hat{x}_t) = \frac{\mathrm{d}\hat{x}_t}{\mathrm{d}t} $  is calculated by taking the time derivative of $\hat{x}_t$. The inference stage is the same as that in standard flow matching.

\subsection{Design Space} 
\label{3.4}
The adjustment term  $\hat{\vb}(\vx_t,t,\vx^r_t)$  and scheduling functions $s(t), \sigma(t)$  constitute the design space of CoFlow.  While $\hat{\vb}(\vx_t,t,\vx^r_t)$ can theoretically take any form, to systematically explore this space, we decouple it into three distinct components:
\begin{equation}
    \hat{\vb}(\vx_t,t,\vx^r_t) = \lambda  \cdot \beta(t) \cdot \vx^{-}.
\end{equation}
\noindent \textbf{Strength Factor $\lambda$}. This hyperparameter controls the overall strength of the repulsive adjustment. 
\begin{wrapfigure}{r}{0.5 \linewidth}
\centering
\vspace{-1em}
\begin{minipage}{0.95\linewidth}
\begin{algorithm}[H]
\caption{CoFlow: Training}
\label{alg:coflow_train}
\begin{lstlisting}
# fn(x, t): velocity network
# x0: training batch of data
t, x1 = rand_t(),randn_like(x0)

# Negative sampling
idx = randperm(x0.shape[0])
x0_r, x1_r = x0[idx], x1[idx]

# Base interpolants and velocities
xt = (1 - t) * x0 + t * x1
xt_r = (1 - t) * x0_r + t * x1_r
vt, vt_r = x1 - x0, x1_r - x0_r

# CoFlow trajectory adjustment
x_hat = xt-lam*t*(1-t)*(xt_r-xt)
v_hat = vt-lam*(1-2*t)*(xt_r-xt)
        -lam*t*(1-t)*(vt_r-vt)

u = fn(x_hat, t)
loss = mse_loss(u, v_hat)
\end{lstlisting}
\end{algorithm}
\end{minipage}
\vspace{-1.5em}
\end{wrapfigure}
\noindent \textbf{Time Schedule $\beta(t)$}. This function dictates how the repulsive adjustment varies across time, providing fine-grained dynamic control. As discussed in Sec~\ref{3.3}, to ensure an unbiased transformation, we enforce the boundary conditions, i.e., $\beta(0) = \beta(1) = 0$. For simplicity, we fix this as a symmetric schedule $\beta(t) = t(1 - t)$.

\noindent \textbf{Negative Signal $\vx^{-}$}. This term defines the specific directional signal used to  repel the current state $\vx_t$ away from the reference trajectory $\vx_t^r$. In this paper, we investigate three distinct formulations for $\vx^-$: 1) \textit{State Residual ($\vx_t^r - \vx_t$)}, the most intuitive spatial signal. Pushing the trajectory in the exact opposite direction of the concurrent negative state at timestep $t$ directly reduces the local Lipschitz constant. 2) \textit{Endpoint Residual ($\vx_1^r - \vx_1$)}. Because the state residual might introduce high variance dynamically across different timesteps, this strategy repels the trajectories at an instance level, thereby reducing temporal variance.  3) \textit{Velocity Residual ($\vv_t^r - \vv_t$)}. Inspired by DeltaFM, this strategy leverages the difference in the target velocities as the directional signal.

\noindent \textbf{Sampling Policy}. This dictates how we obtain the reference trajectory $(\vx_0^r, \vx_1^r)$ within a mini-batch. Specifically, we explore three distinct policies: 1) \textit{Random}: We uniformly sample another trajectory from the current mini-batch to serve as the negative reference. 2) \textit{Nearest}. We search for the most spatially adjacent sample in the batch as the reference. 3) \textit{Largest Lipschitz}. We select the sample that yields the highest local Lipschitz constant relative to the target.

For the remainder of this paper, we default to the linear interpolation configuration, where $s(t) = 1 - t$ and $\sigma(t) = \frac{t}{1-t}$. By default, we adopt the random sampling policy and utilize the state residual $(\vx_t^r - \vx_t)$ as the negative signal. Under this specific setting, the explicit CoFlow stochastic interpolant and its corresponding target velocity field become:
\begin{align}
        \hat{\vx}_t & = (1-t)\vx_0 + t\vx_1 - \lambda t(1-t)(\vx^r_t - \vx_t) \label{eq:coflow_xt}\\
        {\vv}_t(\hat{\vx}_t) & = \vx_1 -\vx_0 - \lambda \big[(1-2t)(\vx^r_t-\vx_t) + t(1-t)(\vv^r_t-\vv_t)\big]
\end{align}
Here, $\vv_t^r = \vx_1^r - \vx_0^r$ and $\vv_t = \vx_1 - \vx_0$ represent the velocities for the reference and target trajectories, respectively. The complete training pipeline is summarized in Algorithm~\ref{alg:coflow_train}.
\section{Experiments}
\begin{figure}[t] 
    \centering
    \includegraphics[width=\linewidth]{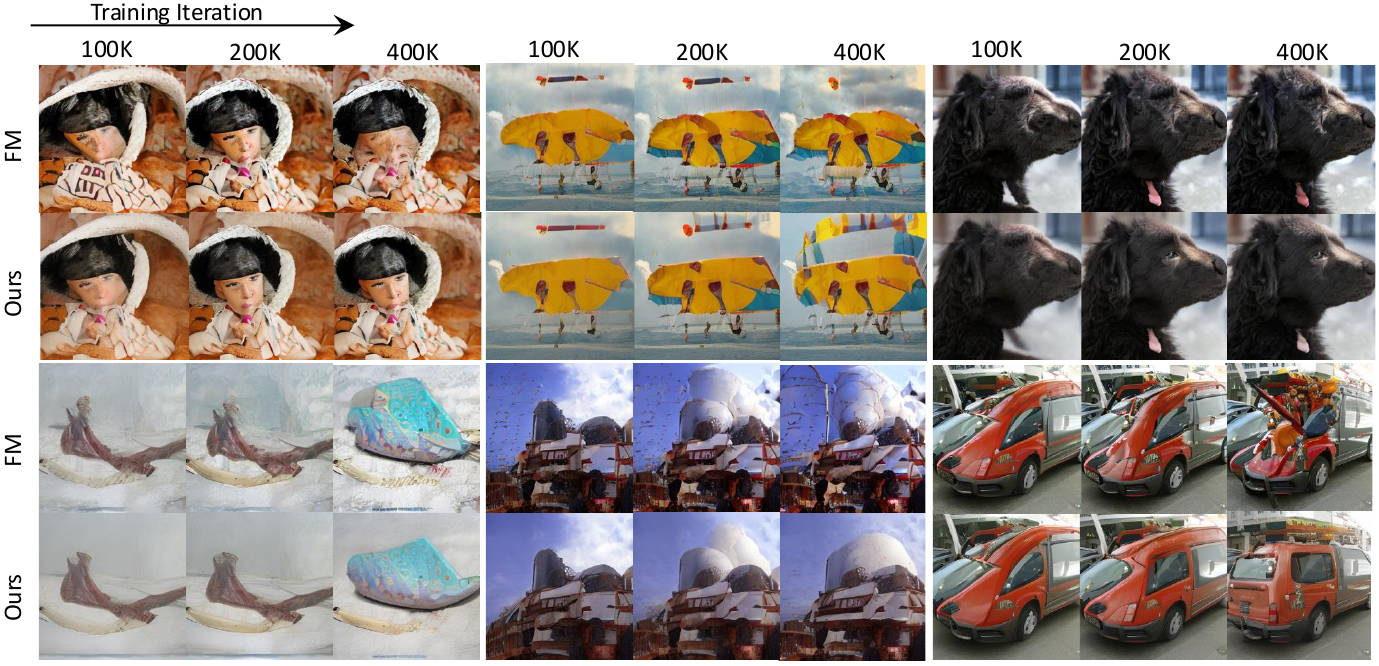}
    \vspace{-1em}
    \caption{ \textbf{Qualitative comparison}. All images are generated using a 20-step EM solver. }
    \label{fig:3}
    \vspace{-1em}
\end{figure}
\noindent \textbf{Experimental Setup}. We evaluated CoFlow on class-conditional ImageNet 256$\times$256 using SiT~\citep{ma2024sit} backbones. Images were encoded into the latent space using a pre-trained SD-VAE~\citep{rombach2022high} before training. Unless otherwise specified, all models were trained for 400K steps with a global batch size of 256, using AdamW~\citep{loshchilov2017decoupled}  with a constant learning rate of $1 \times 10^{-4}$ and no weight decay. For evaluation, we generated 50K images using a 20-step Euler--Maruyama (EM) sampler by default. We reported FID~\citep{heusel2017gans}, sFID~\citep{nash2021generating}, Inception Score (IS)~\citep{salimans2016improved}, Recall~\citep{sajjadi2018assessing}, and Precision~\citep{sajjadi2018assessing}. All remaining settings followed the SiT configuration.

\subsection{Comparison with Standard Flow Matching}

\noindent \textbf{Setting}. We compared CoFlow with standard FM using SiT-B/2 and SiT-XL/2 using the training protocol described above. For each method, we trained with and without REPA. For CoFlow, it was trained using the state residual as the negative signal,  $\lambda = 0.2$, and $\beta(t) = t(1-t)$, as described in  Sec.~\ref{3.4}. The results in Table~\ref{main} are obtained without classifier-free guidance (CFG).

\noindent \textbf{Few-Step Generation}. Table~\ref{main} reports generation quality at NEF$=20$. Without REPA, CoFlow reduces FID from 61.43 to 52.79 on SiT-B/2 and from 38.50 to 29.65 on SiT-XL/2. The other four metrics also improve and the improvements are retained when using REPA.  These results demonstrate the effectiveness of CoFlow in improving few-step generation.
Fig.~\ref{fig:3} provides a qualitative comparison using the 
20-step EM sampler.

\noindent \textbf{Effect of Sampling Budget}. Fig.~\ref{fig:nfe} reports FID-10K  across different numbers of evaluation (NFE). CoFlow substantially outperforms standard FM at low NFEs, and this gap generally narrows as NFE increases, with two methods achieving similar FID at 250 NFEs. This trend is consistent with our motivation:  learning a smoother velocity field with lower local Lipschitz constants can decrease numerical truncation errors under few sampling steps. As the sampling budget increases, finer discretization reduces these errors for both methods, diminishing the advantage of CoFlow. These results suggest that CoFlow primarily improves sampling efficiency in the few-step setting.

\begin{table}[t]
    \centering
    \caption{   \textbf{Class-Conditional ImageNet 256$\times$256.} Training for 400k steps. We sample 50K images using the EMSolver, with the number of function evaluations (NFE) set to 20. No CFG involved.}
    \label{main}
    
    \renewcommand{\arraystretch}{1.1} 
    \setlength{\tabcolsep}{11pt}  
   {
    \begin{tabular}{llccccc}
        \toprule
        Model & Method & FID$\downarrow$ & sFID$\downarrow$ & IS$\uparrow$ & Precision$\uparrow$ & Recall$\uparrow$ \\
        \midrule
        
        \multirow{4}{*}{SiT-B/2}  & FM  & 61.43 & 35.18& 27.46& 0.43 & {0.54} \\
                                  & \cellcolor{coflowdefault}CoFlow & \cellcolor{coflowdefault}\textbf{52.79} & \cellcolor{coflowdefault}\textbf{21.71}& \cellcolor{coflowdefault}\textbf{35.00}&\cellcolor{coflowdefault}\textbf{0.45} & \cellcolor{coflowdefault}\textbf{0.58} \\
        \cmidrule(l){2-7}
                                  & FM + REPA  & 47.64 &34.33 & 40.75 & 0.48 & {0.59} \\
                                  & \cellcolor{coflowdefault}CoFlow + REPA & \cellcolor{coflowdefault}\textbf{37.76}& \cellcolor{coflowdefault}\textbf{20.23}& \cellcolor{coflowdefault}\textbf{52.92}& \cellcolor{coflowdefault}\textbf{0.51}& \cellcolor{coflowdefault}\textbf{0.60}\\
        
        \midrule
        
        \multirow{4}{*}{SiT-XL/2} 
                                  & FM   & 38.50 & 29.27 & 50.14 & 0.54 & {0.58} \\
                                  & \cellcolor{coflowdefault}CoFlow& \cellcolor{coflowdefault}\textbf{29.65}& \cellcolor{coflowdefault}\textbf{17.41}&\cellcolor{coflowdefault}\textbf{64.12} &\cellcolor{coflowdefault}\textbf{0.57} &\cellcolor{coflowdefault}\textbf{0.59} \\
            \cmidrule(l){2-7}
                                & FM + REPA  & 35.35 &27.98 &63.13 & 0.52& \textbf{0.55} \\
                                  & \cellcolor{coflowdefault}CoFlow+ REPA & \cellcolor{coflowdefault}\textbf{27.78}& \cellcolor{coflowdefault}\textbf{17.88}&\cellcolor{coflowdefault}\textbf{80.52} &\cellcolor{coflowdefault}\textbf{0.55} &\cellcolor{coflowdefault}0.53 \\
        \bottomrule
    \end{tabular}
    }
\end{table}

\begin{figure*}[t]
    \centering

    \begin{minipage}[t]{0.48\textwidth}
        \vspace{0pt}
        \centering
        \setlength{\tabcolsep}{5pt}
        \renewcommand{\arraystretch}{1.1}

        \captionof{table}{
             \textbf{Compatibility with DeltaFM.} Trained on SiT-B/2 + REPA for 400K steps.
        }
        \label{tab:compatibility-deltafm}
        \vspace{-1em}
        \begin{tabular*}{\linewidth}{
            @{\extracolsep{\fill}}lrrr@{}
        }
            \toprule
            Method & FID$\downarrow$ & sFID$\downarrow$ & IS$\uparrow$ \\
            \midrule
            FM         & 47.64 &34.33 & 40.75  \\
            DeltaFM    & 32.32 & 15.73 & 56.33 \\
            CoFlow     & 37.76 & 20.23 & 52.92 \\
            CoFlow+VC  & \textbf{27.74}
                       & \textbf{8.68}
                       & \textbf{63.94} \\
            \bottomrule
        \end{tabular*}

        \vspace{0.2em}

        \captionof{table}{  
       \textbf{Compatibility with OT-CFM.} Trained on SiT-B/2 without REPA.
       }
        \label{tab:compatibility-otcfm}

        \begin{tabular*}{\linewidth}{
            @{\extracolsep{\fill}}lrrr@{}
        }
            \toprule
            Method & FID$\downarrow$ & sFID$\downarrow$ & IS$\uparrow$ \\
            \midrule
            OT-CFM        & 69.90 & 38.87 & 23.47 \\
            \hspace{1em}+CoFlow & \textbf{63.92}
                          & \textbf{25.96}
                          & \textbf{28.75} \\
            \bottomrule
        \end{tabular*}
    \end{minipage}
    \hfill
    \begin{minipage}[t]{0.49\textwidth}
        \vspace{0.8em}
        \centering

        \includegraphics[width=\linewidth]{
            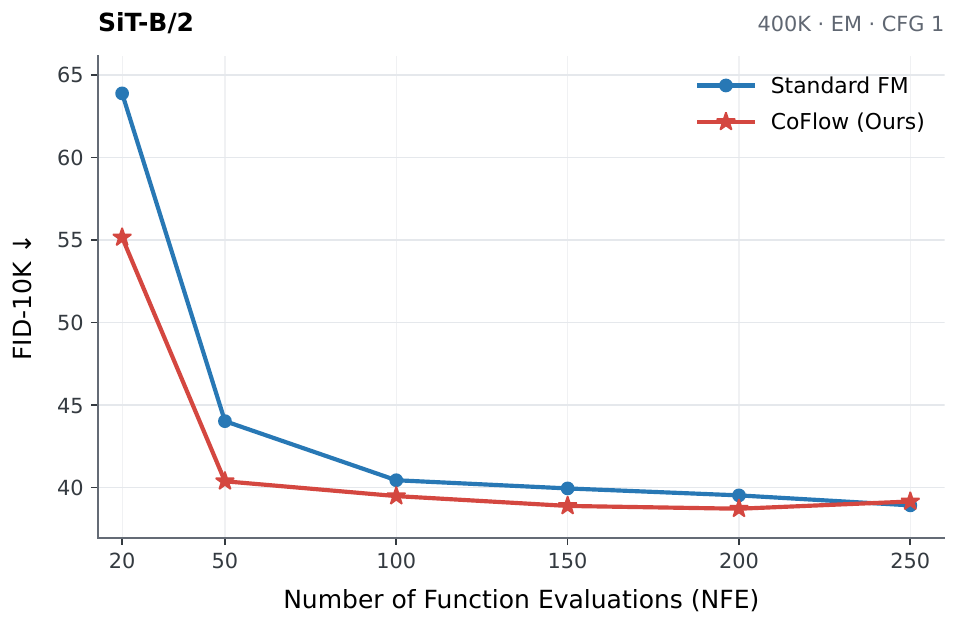}
        \captionof{figure}{ 
           \textbf{FID-10K versus NFE for Standard FM and CoFlow.} EM solver was used to sample on SiT-B/2 without REPA.
        }
        \label{fig:nfe}
    \end{minipage}
\vspace{-1em}
\end{figure*}

\subsection{Compatibility with Other Methods}
We investigated CoFlow's compatibility with two closely related methods: DeltaFM, which also exploits negative samples, and OT-CFM, which seeks to mitigate trajectory
crossing.

\noindent \textbf {DeltaFM.} DeltaFM incorporates negative samples through modifying the 
training objective, whereas CoFlow adjusts the intermediate trajectories. Although DeltaFM was originally introduced as a training-based method, we show that its optimal velocity field can be obtained by an affine transformation of the optimal
standard FM field. We therefore apply this transformation to a trained
standard FM model to approximate the DeltaFM solution without additional training:
\begin{equation}
    \vu_\theta^{\mathrm{\Delta FM}}(\vx_t)
    =(\vu_\theta^{\mathrm{FM}}(\vx_t)
        -\lambda\vm_t) /  (1-\lambda),
    \label{eq:vc}
\end{equation}
where $\vm_t$ is calculated by averaging the  latents across the entire
training set. We refer to this equivalent training-free operation as velocity centering (VC) and provide the derivation in Appendix~\ref{AppendixD}.
As shown in Table~\ref{tab:deltafm}, applying VC to
trained FM models yields performance comparable to
models trained with the DeltaFM objective.

We next evaluated whether VC could further improve CoFlow, using SiT-B/2 trained for 400K steps with REPA. As shown in Table~\ref{tab:compatibility-deltafm},
applying VC reduces CoFlow's FID from 37.76 to 27.74 and outperforms DeltaFM on all three metrics. These results demonstrate that CoFlow's trajectory
adjustment can be effectively combined with DeltaFM for further improvement.

\noindent \textbf{OT-CFM.} OT-CFM attempts to reduce trajectory crossing by optimizing the coupling between noise and data, whereas CoFlow directly adjusts the
intermediate trajectories. To evaluate whether they are compatible, we combined CoFlow with OT-CFM using SiT-B/2 trained for 400K steps
without REPA. Table~\ref{tab:compatibility-otcfm} shows that incorporating CoFlow with OT-CFM reduces FID from 69.90 to 63.92 and sFID
from 38.87 to 25.96. These improvements indicate that trajectory adjustment remains beneficial when combined with optimized endpoint coupling.

\begin{table}[t]
    \centering
    \caption{
        \textbf{Component-wise ablation on ImageNet
        256$\times$256.}
        All models were SiT-S/2 trained for 400K iterations
        without REPA.
        We evaluated 50K images using the EM
        sampler with NFE $=20$ and fixed
        $\beta(t)=t(1-t)$.
        Colored cells indicate the factor varied in each block.
    }
    \label{tab:coflow_ablation}
    \setlength{\tabcolsep}{6pt}
    \renewcommand{\arraystretch}{1.1}

    \begin{tabularx}{\linewidth}{
        c
        >{\centering\arraybackslash}X
        >{\centering\arraybackslash}X
        r r r
    }
        \toprule
        $\lambda$
        & Negative signal $x^{-}$
        & Sampling policy
        & FID$\downarrow$
        & sFID$\downarrow$
        & IS$\uparrow$ \\
        \midrule

        \multicolumn{3}{l}{Standard FM}
        & 86.99 & 43.47 & 16.25 \\
        \midrule

        $0.2$
        & \cellcolor{ablSignal}$\epsilon$
        & Random
        & 86.49 & 41.77 & 16.49 \\

        $0.2$
        & \cellcolor{ablSignal}$x_t^r-x_t$
        & Random
        & {79.55}
        & {26.63}
        & {20.36} \\

        $0.2$
        & \cellcolor{ablSignal}$x_1^r-x_1$
        & Random
        & 92.24 & 47.22 & 16.25 \\

        $0.2$
        & \cellcolor{ablSignal}$v_t^r-v_t$
        & Random
        & 122.19 & 87.04 & 9.41 \\
        \midrule

        $0.2$
        & $x_t^r-x_t$
        & \cellcolor{ablPolicy}Random
        & 79.55 & 26.63 & 20.36 \\

        $0.2$
        & $x_t^r-x_t$
        & \cellcolor{ablPolicy}Lipschitz
        & {78.12}
        & 26.16
        & {20.66} \\

        $0.2$
        & $x_t^r-x_t$
        & \cellcolor{ablPolicy}Nearest
        & 79.40
        & {24.64}
        & 20.27 \\
        \midrule

        \cellcolor{ablStrength}$0.05$
        & $x_t^r-x_t$
        & Random
        & 85.66 & 39.77 & 17.04 \\

        \cellcolor{ablStrength}$0.2$
        & $x_t^r-x_t$
        & Random
        & 79.55 & 26.63 & 20.36 \\

        \cellcolor{ablStrength}$0.5$
        & $x_t^r-x_t$
        & Random
        & {75.06}
        & {14.82}
        & {22.45} \\

        \cellcolor{ablStrength}$1.0$
        & $x_t^r-x_t$
        & Random
        & 87.43 & 56.73 & 14.15 \\
        \bottomrule
    \end{tabularx}

\end{table}

\subsection{Component-Wise Ablation}

We examined the negative signal, sampling policy, and
repulsive strength using SiT-S/2 trained for 400K steps
without REPA.
Table~\ref{tab:coflow_ablation} reports the results of
varying one component at a time.
The default configuration uses $\lambda=0.2$, the state
residual $(\vx_t^r-\vx_t)$, and random negative sampling.

\noindent
\ablationheading{ablSignalTitle}{Negative signal.}\enspace
We first used random noise as signal to test whether random perturbations alone could explain CoFlow's gains. This control changes FID only slightly, from 86.99 to 86.49, suggesting that simply perturbing the trajectories cannot bring the observed improvement. We then compared three signals derived from negative
trajectories: the state, endpoint, and velocity residuals.
The state residual $(\vx_t^r-\vx_t)$ performs best across
all three metrics, achieving an FID of 79.55.
We hypothesize that the state residual is more suitable
to reduce local Lipschitz constants, as it guides
repulsion using the relative positions of trajectories.

\noindent
\ablationheading{ablPolicyTitle}{Sampling policy.}\enspace
We compared the Random, Nearest, and Lipschitz policies described in Sec.~\ref{3.4}. Although negatives can be selected using states at time $t$, the selected negatives may change as $t$ varies, introducing discontinuities in the stochastic interpolant in Eq.~(\ref{eq:coflow_xt}).
To avoid this, we used endpoint states $\vx_0$ to select
negatives for the Nearest and Lipschitz policies as a
practical compromise. It turns out all three policies improve upon standard FM, with
Lipschitz achieving the lowest FID of 78.12. This is consistent with our motivation to improve few-step generation by reducing local Lipschitz constants. Because random sampling remains competitive with an FID of 79.55, we adopt it as the default policy for its simplicity.

\noindent
\ablationheading{ablStrengthTitle}{Strength factor.}\enspace
We varied $\lambda\in\{0.05,0.2,0.5,1.0\}$, with
$\lambda=0$ corresponding to standard FM.
Increasing $\lambda$ from 0.05 to 0.5 improves FID
from 85.66 to 75.06. However, increasing $\lambda$ further to 1.0 degrades
FID to 87.43, slightly worse than standard FM.
These results show that moderate repulsion improves
generation quality, while an excessively strong
adjustment can be detrimental.

\section{Conclusion}
This paper presents CoFlow, a framework  that utilizes negative samples to repel crossing training trajectories, which is inspired by Doob-\textit{h} transform. Motivated by the relationship between trajectory crossing and local Lipschitze constants, CoFlow adjusts intermediate to encourage a smoother velocity field. We derived a closed-form stochastic interpolant that preserves the endpoint distributions and enables simulation-free training. Experiments on ImageNet $256\times256$ demonstrate improved few-step generation and compatibility with
methods like REPA and OT-CFM. We also developed an inference-time velocity adjustment that is equivalent to DeltaFM, which can further improves CoFlow without additional training. Our findings suggest that contrastive learning can serve as a mechanism for adjusting training trajectories to improve few-step generation in flow matching.

\section*{AI Use Statement}
In this work, we used generative AI tools (specifically, ChatGPT) solely for assistance with language polishing and grammatical improvements. We have not used generative AI tools for research ideation, data collection, mathematical proofs, experimental implementation, or writing the scientific claims of this paper. All AI-assisted revisions were thoroughly reviewed and verified by the authors. We take full responsibility for the final content of this work, including all text, claims, and artifacts.

\section*{Reproducibility Statement}
We provide the CoFlow training procedure in Algorithm~\ref{alg:coflow_train} and describe its design choices and default configuration in Sec.~\ref{3.4}. The training and evaluation settings, together with component-wise ablations, are reported in Sec.~4. Supporting derivations and assumptions are provided in Appendices~\ref{appendixA}--\ref{AppendixD}, including the formulation and implementation of velocity centering. Our anonymous implementation is available at \url{https://anonymous.4open.science/r/CoFlow-A0A0}.

\bibliographystyle{iclr2027_conference}
\bibliography{iclr2027_conference}

@article{liu2022flow,
  title={Flow straight and fast: Learning to generate and transfer data with rectified flow},
  author={Liu, Xingchao and Gong, Chengyue and Liu, Qiang},
  journal={arXiv preprint arXiv:2209.03003},
  year={2022}
}

@article{lipman2022flow,
  title={Flow matching for generative modeling},
  author={Lipman, Yaron and Chen, Ricky TQ and Ben-Hamu, Heli and Nickel, Maximilian and Le, Matt},
  journal={arXiv preprint arXiv:2210.02747},
  year={2022}
}

@article{ho2020denoising,
  title={Denoising diffusion probabilistic models},
  author={Ho, Jonathan and Jain, Ajay and Abbeel, Pieter},
  journal={Advances in neural information processing systems},
  volume={33},
  pages={6840--6851},
  year={2020}
}

@article{karras2022elucidating,
  title={Elucidating the design space of diffusion-based generative models},
  author={Karras, Tero and Aittala, Miika and Aila, Timo and Laine, Samuli},
  journal={Advances in neural information processing systems},
  volume={35},
  pages={26565--26577},
  year={2022}
}

@article{albergo2022building,
  title={Building normalizing flows with stochastic interpolants},
  author={Albergo, Michael S and Vanden-Eijnden, Eric},
  journal={arXiv preprint arXiv:2209.15571},
  year={2022}
}

@inproceedings{rombach2022high,
  title={High-resolution image synthesis with latent diffusion models},
  author={Rombach, Robin and Blattmann, Andreas and Lorenz, Dominik and Esser, Patrick and Ommer, Bj{\"o}rn},
  booktitle={Proceedings of the IEEE/CVF conference on computer vision and pattern recognition},
  pages={10684--10695},
  year={2022}
}

@article{song2020score,
  title={Score-based generative modeling through stochastic differential equations},
  author={Song, Yang and Sohl-Dickstein, Jascha and Kingma, Diederik P and Kumar, Abhishek and Ermon, Stefano and Poole, Ben},
  journal={arXiv preprint arXiv:2011.13456},
  year={2020}
}

@inproceedings{ma2024sit,
  title={Sit: Exploring flow and diffusion-based generative models with scalable interpolant transformers},
  author={Ma, Nanye and Goldstein, Mark and Albergo, Michael S and Boffi, Nicholas M and Vanden-Eijnden, Eric and Xie, Saining},
  booktitle={European Conference on Computer Vision},
  pages={23--40},
  year={2024},
  organization={Springer}
}

@article{yu2024representation,
  title={Representation alignment for generation: Training diffusion transformers is easier than you think},
  author={Yu, Sihyun and Kwak, Sangkyung and Jang, Huiwon and Jeong, Jongheon and Huang, Jonathan and Shin, Jinwoo and Xie, Saining},
  journal={arXiv preprint arXiv:2410.06940},
  year={2024}
}

@article{goodfellow2014generative,
  title={Generative adversarial nets},
  author={Goodfellow, Ian J and Pouget-Abadie, Jean and Mirza, Mehdi and Xu, Bing and Warde-Farley, David and Ozair, Sherjil and Courville, Aaron and Bengio, Yoshua},
  journal={Advances in neural information processing systems},
  volume={27},
  year={2014}
}

@inproceedings{yin2024one,
  title={One-step diffusion with distribution matching distillation},
  author={Yin, Tianwei and Gharbi, Micha{\"e}l and Zhang, Richard and Shechtman, Eli and Durand, Fredo and Freeman, William T and Park, Taesung},
  booktitle={Proceedings of the IEEE/CVF conference on computer vision and pattern recognition},
  pages={6613--6623},
  year={2024}
}

@article{salimans2022progressive,
  title={Progressive distillation for fast sampling of diffusion models},
  author={Salimans, Tim and Ho, Jonathan},
  journal={arXiv preprint arXiv:2202.00512},
  year={2022}
}

@inproceedings{song2023consistency,
  title={Consistency models},
  author={Song, Yang and Dhariwal, Prafulla and Chen, Mark and Sutskever, Ilya},
  booktitle={International conference on machine learning},
  year={2023}
}

@inproceedings{rahaman2019spectral,
  title={On the spectral bias of neural networks},
  author={Rahaman, Nasim and Baratin, Aristide and Arpit, Devansh and Draxler, Felix and Lin, Min and Hamprecht, Fred and Bengio, Yoshua and Courville, Aaron},
  booktitle={International conference on machine learning},
  pages={5301--5310},
  year={2019},
  organization={PMLR}
}

@inproceedings{chen2020simple,
  title={A simple framework for contrastive learning of visual representations},
  author={Chen, Ting and Kornblith, Simon and Norouzi, Mohammad and Hinton, Geoffrey},
  booktitle={International conference on machine learning},
  pages={1597--1607},
  year={2020},
  organization={PmLR}
}

@inproceedings{he2020momentum,
  title={Momentum contrast for unsupervised visual representation learning},
  author={He, Kaiming and Fan, Haoqi and Wu, Yuxin and Xie, Saining and Girshick, Ross},
  booktitle={Proceedings of the IEEE/CVF conference on computer vision and pattern recognition},
  pages={9729--9738},
  year={2020}
}

@inproceedings{stoica2025contrastive,
  title={Contrastive flow matching},
  author={Stoica, George and Ramanujan, Vivek and Fan, Xiang and Farhadi, Ali and Krishna, Ranjay and Hoffman, Judy},
  booktitle={Proceedings of the IEEE/CVF International Conference on Computer Vision},
  pages={1185--1194},
  year={2025}
}

@article{ho2022classifier,
  title={Classifier-free diffusion guidance},
  author={Ho, Jonathan and Salimans, Tim},
  journal={arXiv preprint arXiv:2207.12598},
  year={2022}
}

@inproceedings{deng2009imagenet,
  title={Imagenet: A large-scale hierarchical image database},
  author={Deng, Jia and Dong, Wei and Socher, Richard and Li, Li-Jia and Li, Kai and Fei-Fei, Li},
  booktitle={2009 IEEE conference on computer vision and pattern recognition},
  pages={248--255},
  year={2009},
  organization={Ieee}
}

@article{tong2023improving,
  title={Improving and generalizing flow-based generative models with minibatch optimal transport},
  author={Tong, Alexander and Fatras, Kilian and Malkin, Nikolay and Huguet, Guillaume and Zhang, Yanlei and Rector-Brooks, Jarrid and Wolf, Guy and Bengio, Yoshua},
  journal={arXiv preprint arXiv:2302.00482},
  year={2023}
}

@article{song2019generative,
  title={Generative modeling by estimating gradients of the data distribution},
  author={Song, Yang and Ermon, Stefano},
  journal={Advances in neural information processing systems},
  volume={32},
  year={2019}
}

@article{dhariwal2021diffusion,
  title={Diffusion models beat gans on image synthesis},
  author={Dhariwal, Prafulla and Nichol, Alexander},
  journal={Advances in neural information processing systems},
  volume={34},
  pages={8780--8794},
  year={2021}
}

@inproceedings{liu2023instaflow,
  title={Instaflow: One step is enough for high-quality diffusion-based text-to-image generation},
  author={Liu, Xingchao and Zhang, Xiwen and Ma, Jianzhu and Peng, Jian and others},
  booktitle={The Twelfth International Conference on Learning Representations},
  year={2023}
}

@article{geng2025mean,
  title={Mean flows for one-step generative modeling},
  author={Geng, Zhengyang and Deng, Mingyang and Bai, Xingjian and Kolter, J Zico and He, Kaiming},
  journal={arXiv preprint arXiv:2505.13447},
  year={2025}
}

@article{frans2024one,
  title={One step diffusion via shortcut models},
  author={Frans, Kevin and Hafner, Danijar and Levine, Sergey and Abbeel, Pieter},
  journal={arXiv preprint arXiv:2410.12557},
  year={2024}
}

@article{deng2026generative,
  title={Generative Modeling via Drifting},
  author={Deng, Mingyang and Li, He and Li, Tianhong and Du, Yilun and He, Kaiming},
  journal={arXiv preprint arXiv:2602.04770},
  year={2026}
}

@article{albergo2023stochastic,
  title={Stochastic interpolants with data-dependent couplings},
  author={Albergo, Michael S and Goldstein, Mark and Boffi, Nicholas M and Ranganath, Rajesh and Vanden-Eijnden, Eric},
  journal={arXiv preprint arXiv:2310.03725},
  year={2023}
}

@article{zhou2023denoising,
  title={Denoising diffusion bridge models},
  author={Zhou, Linqi and Lou, Aaron and Khanna, Samar and Ermon, Stefano},
  journal={arXiv preprint arXiv:2309.16948},
  year={2023}
}

@article{chung2022diffusion,
  title={Diffusion posterior sampling for general noisy inverse problems},
  author={Chung, Hyungjin and Kim, Jeongsol and Mccann, Michael T and Klasky, Marc L and Ye, Jong Chul},
  journal={arXiv preprint arXiv:2209.14687},
  year={2022}
}

@article{liu20232,
  title={{I$^2$SB}: Image-to-image Schr{\"o}dinger bridge},
  author={Liu, Guan-Horng and Vahdat, Arash and Huang, De-An and Theodorou, Evangelos A and Nie, Weili and Anandkumar, Anima},
  journal={arXiv preprint arXiv:2302.05872},
  year={2023}
}

@article{kawar2022denoising,
  title={Denoising diffusion restoration models},
  author={Kawar, Bahjat and Elad, Michael and Ermon, Stefano and Song, Jiaming},
  journal={Advances in neural information processing systems},
  volume={35},
  pages={23593--23606},
  year={2022}
}

@article{heusel2017gans,
  title={Gans trained by a two time-scale update rule converge to a local nash equilibrium},
  author={Heusel, Martin and Ramsauer, Hubert and Unterthiner, Thomas and Nessler, Bernhard and Hochreiter, Sepp},
  journal={Advances in neural information processing systems},
  volume={30},
  year={2017}
}

@article{nash2021generating,
  title={Generating images with sparse representations},
  author={Nash, Charlie and Menick, Jacob and Dieleman, Sander and Battaglia, Peter W},
  journal={arXiv preprint arXiv:2103.03841},
  year={2021}
}

@article{salimans2016improved,
  title={Improved techniques for training gans},
  author={Salimans, Tim and Goodfellow, Ian and Zaremba, Wojciech and Cheung, Vicki and Radford, Alec and Chen, Xi},
  journal={Advances in neural information processing systems},
  volume={29},
  year={2016}
}

@article{sajjadi2018assessing,
  title={Assessing generative models via precision and recall},
  author={Sajjadi, Mehdi SM and Bachem, Olivier and Lucic, Mario and Bousquet, Olivier and Gelly, Sylvain},
  journal={Advances in neural information processing systems},
  volume={31},
  year={2018}
}

@article{davtyan2026faster,
  title={Faster inference of flow-based generative models via improved data-noise coupling},
  author={Davtyan, Aram and Dadi, Leello Tadesse and Cevher, Volkan and Favaro, Paolo},
  journal={arXiv preprint arXiv:2603.15279},
  year={2026}
}

@article{loshchilov2017decoupled,
  title={Decoupled weight decay regularization},
  author={Loshchilov, Ilya and Hutter, Frank},
  journal={arXiv preprint arXiv:1711.05101},
  year={2017}
}

\clearpage
\appendix

\section{Limitations}
Our image-generation experiments focus on class-conditional ImageNet 256 $\times$256 with SiT backbones, due to limited computational resources. The effectiveness of CoFlow on other datasets, higher resolutions, and different generation tasks remains to be evaluated. In addition, our current implementation uses a fixed repulsion schedule and selects negative samples independently of time to retain a simple closed-form target velocity. Future work could explore adaptive schedules and time-dependent negative selection while preserving smooth interpolants and tractable velocity targets.

\section{Marginal Equivalence between SDE and SI}
\label{appendixA}
In this section, we give a self-contained proof of related conclusions in  Sec.~\ref{3.1}. Formally, we need to for  the following SDE:
\begin{equation}
    \label{eq13}
    \mathrm{d}\vx_t = \vf(\vx_t,t) \, \mathrm{d}t + g(t) \, \mathrm{d}\vw_t,
\end{equation}
it admits the following representation in distribution for each fixed time $t$:
\begin{equation}
    \vx_t  \overset{d}{=} s(t)\vx_0 +  s(t)\sigma(t)\vx_1, 
\end{equation}
where $\vf(\vx_t,t) = \alpha(t) \vx_t$, and $\overset{d}{=}$ denotes equality in distribution.. A typical method is to introduce an integrating factor as follow:
\begin{equation}
    s(t) = \exp\left(\int_0^t \alpha(\xi) d\xi\right)
\end{equation}
Take the derivative with respect to time \(t\) of $s(t)$, we have:
\begin{equation}
    \frac{\mathrm{d}s(t)}{\mathrm{d}t} = \alpha(t) \cdot \exp\left(\int_0^t \alpha(\xi) d\xi\right) = \alpha(t)s(t)
\end{equation}
Which means: 
\begin{equation}
    \mathrm{d}s(t) = \alpha(t)s(t) \mathrm{d}t
\end{equation}
Then we consider the differential of $\frac{\vx_t}{s(t)}$:
\begin{align}
    \mathrm{d}\left(\frac{{\vx}_t}{s(t)}\right) &= \frac{\mathrm{d}{\vx}_t}{s(t)} - \frac{{\vx}_t \mathrm{d}s(t)}{s(t)^2} \\
    \mathrm{d}\left(\frac{{\vx}_t}{s(t)}\right) &= \frac{\alpha(t){\vx}_t \mathrm{d}t + g(t) \mathrm{d}\vw_t}{s(t)} - \frac{{\vx}_t (\alpha(t)s(t) \mathrm{d}t)}{s(t)^2} \\
    \mathrm{d}\left(\frac{{\vx}_t}{s(t)}\right) &= \left[ \frac{\alpha(t){\vx}_t}{s(t)} \mathrm{d}t + \frac{g(t)}{s(t)} \mathrm{d}\vw_t \right] - \frac{\alpha(t){\vx}_t}{s(t)} \mathrm{d}t \\
     \mathrm{d}\left(\frac{{\vx}_t}{s(t)}\right) &=  \frac{g(t)}{s(t)} \mathrm{d}\vw_t  
\end{align}
Then we can integrate both sides of the equation from $0$ to $t$ to solve $x_t$:
\begin{align}
    \int_0^t \mathrm{d}\left(\frac{{\vx}_\xi}{s(\xi)}\right) &= \int_0^t \frac{g(\xi)}{s(\xi)} \mathrm{d}\vw_\xi \\
    \frac{{\vx}(t)}{s(t)} - \frac{{\vx}(0)}{s(0)} &= \int_0^t \frac{g(\xi)}{s(\xi)} \mathrm{d}\vw_\xi 
\end{align}
According to the definition, $s(0) = \exp\left(\int_0^0 \alpha(\xi) d\xi\right) = \exp(0) = 1$, therefore we have:
\begin{equation}
    {\vx}(t) = s(t){\vx}(0) + s(t)\int_0^t \frac{g(\xi)}{s(\xi)} \mathrm{d}\vw_\xi 
\end{equation}
According to the properties of stochastic calculus, the integral of a deterministic function $\frac{g(\xi)}{s(\xi)}$ with respect to Brownian motion $d\vw_\xi$ follows a normal distribution with a mean of 0 and a variance equal to the integral of the square of that function (known as the Itô Isometry):
\begin{equation}
    \int_0^t \frac{g(\xi)}{s(\xi)} d\vw_\xi \sim \mathcal{N}\left(0, \left( \int_0^t \frac{g(\xi)^2}{s(\xi)^2} d\xi \right)\mI\right).
\end{equation}
Therefore, let $\sigma(t)^2 = \int_0^t \frac{g(\xi)^2}{s(\xi)^2} d\xi$, 
and $\vx_1 \sim \mathcal{N}(0,\mI)$, the solution can be reparameterized as Eq~(\ref{eq2}):
\begin{align}
        \vx_t & \overset{d}{=} s(t)\vx_0 +  s(t)\sigma(t)\vx_1 \\
        s(t) &= \exp\left(\int_0^t \alpha(\xi) d\xi\right) \\
        \sigma(t)^2 &= \int_0^t \frac{g(\xi)^2}{s(\xi)^2} d\xi 
\end{align}
The results are consistent with those in EDM~\citep{karras2022elucidating}. This representation matches the SDE’s one-time marginals; using a fixed $\vx_1$ across time defines an interpolant that need not have the same joint distribution across time.
For the special configuration, where  $\alpha(t) = \frac{-1}{1-t}$ and $g(t) = \sqrt{\frac{2t}{1-t}}$, substitute them into the definition of $s(t), \sigma(t)$ :
\begin{equation}
\label{eq29}
     s(t) = \exp\left(\int_0^t  \frac{-1}{1-\xi} d\xi\right)  = 1 - t.
\end{equation}
\begin{equation}
\label{eq30}
       \sigma(t)^2 = \int_0^t \frac{2 \xi}{(1-\xi)^3} d\xi = \frac{t^2}{(1-t)^2}
\end{equation}
Therefore, $\vx_t  \overset{d}{=}  s(t)\vx_0 + s(t)\sigma(t)\vx_1 = (1-t)\vx_0 + t\vx_1$, which is linear.

\section{Equivalence of CoFlow}
\label{AppendixB}
In this section, we prove the equivalence between Eq~(\ref{eq5}) and Eq(~\ref{eq6}). Formally, for the following SDE:
\begin{equation}
     \mathrm{d}\vx_t = \left [ \vf(\vx_t,t) - \vb(\vx_t,t, \vx^r_t ) \right ]\, \mathrm{d}t + g(t) \, \mathrm{d}\vw_t
\end{equation}
Its solution should be in the form as follows:
\begin{equation}
    \hat{\vx}_t = s(t)\vx_0 + s(t)\sigma(t)\vx_1 - s(t)\int_0^t\frac{\vb(\vx_\xi,\xi,\vx^r_\xi)}{s(\xi)}\mathrm{d}\xi,
\end{equation}
The proof is similar to that in Appendix~\ref{appendixA}, with the same definition of $s(t)$ and $\sigma(t)$. Specifically, while adding an additional drift $\vb(x_t,t,x^r_t)$ to Eq~(\ref{eq1}), the differential of $\frac{\vx_t}{s(t)}$ becomes:

\begin{align}
    \mathrm{d}\left(\frac{{\vx}_t}{s(t)}\right) &= \frac{\mathrm{d}{\vx}_t}{s(t)} - \frac{{\vx}_t \mathrm{d}s(t)}{s(t)^2} \\
    \mathrm{d}\left(\frac{{\vx}_t}{s(t)}\right) &= \frac{\alpha(t){\vx}_t \mathrm{d}t   \textcolor{red}{- \vb(\vx_t,t, \vx^r_t )\mathrm{d}t} + g(t) \mathrm{d}\vw_t}{s(t)} - \frac{{\vx}_t (\alpha(t)s(t) \mathrm{d}t)}{s(t)^2} \\
    \mathrm{d}\left(\frac{{\vx}_t}{s(t)}\right) &= \left[ \frac{\alpha(t){\vx}_t}{s(t)} \mathrm{d}t  \textcolor{red}{- \frac{\vb(\vx_t,t, \vx^r_t )}{s(t)}\mathrm{d}t} + \frac{g(t)}{s(t)} \mathrm{d}\vw_t \right] - \frac{\alpha(t){\vx}_t}{s(t)} \mathrm{d}t \\
     \mathrm{d}\left(\frac{{\vx}_t}{s(t)}\right) &=  \frac{g(t)}{s(t)} \mathrm{d}\vw_t  \textcolor{red}{- \frac{\vb(\vx_t,t, \vx^r_t )}{s(t)}\mathrm{d}t}
\end{align}

Similarly, we integrate both sides of the equation from $0$ to $t$ to solve $x_t$:
\begin{align}
    \int_0^t \mathrm{d}\left(\frac{{\vx}_\xi}{s(\xi)}\right) &= \int_0^t \frac{g(\xi)}{s(\xi)} \mathrm{d}\vw_\xi \textcolor{red}{- \int^t_0\frac{\vb(\vx_\xi,\xi, \vx^r_\xi )}{s(\xi)}\mathrm{d}\xi} \\
    \frac{{\vx}(t)}{s(t)} - \frac{{\vx}(0)}{s(0)} &= \int_0^t \frac{g(\xi)}{s(\xi)} \mathrm{d}\vw_\xi \textcolor{red}{- \int^t_0\frac{\vb(\vx_\xi,\xi, \vx^r_\xi )}{s(\xi)}\mathrm{d}\xi}
\end{align}
Therefore, we can solve $x_t$:
\begin{equation}
    {\vx}(t) = s(t){\vx}(0) + s(t)\int_0^t \frac{g(\xi)}{s(\xi)} \mathrm{d}\vw_\xi \textcolor{red}{-s(t) \int^t_0\frac{\vb(\vx_\xi,\xi, \vx^r_\xi )}{s(\xi)}\mathrm{d}\xi}
\end{equation}
Because $\vf(\vx_t,t), s(t), \sigma(t), g(t)$ are identical in the two SDEs(Eq~(\ref{eq1}) and Eq~(\ref{eq5}))), $\vx_t$ can be simplified into:
\begin{equation}
    \hat{\vx}_t = s(t)\vx_0 + s(t)\sigma(t)\vx_1 \textcolor{red}{- s(t)\int_0^t\frac{\vb(\vx_\xi,\xi,\vx^r_\xi)}{s(\xi)}\mathrm{d}\xi}
\end{equation}

Additionally, for the \textbf{Simulation-Free Reparameterization}, we define $\hat{\vb}(\vx_t,t,\vx^r_t)$ as:
\begin{align}
\hat{\vb}(\vx_t,t,\vx^r_t) &= s(t)\int_0^t\frac{\vb(\vx_\xi,\xi,\vx^r_\xi)}{s(\xi)}\mathrm{d}\xi \\
\frac{\hat{\vb}(\vx_t,t,\vx^r_t) }{s(t)}&= \int_0^t\frac{\vb(\vx_\xi,\xi,\vx^r_\xi)}{s(\xi)}\mathrm{d}\xi 
\end{align}
Take the time derivative of both sides:
\begin{align}
\frac{\mathrm{d}}{\mathrm{d}t} \big[ \hat{\vb}(\vx_t, t, \vx_t^r) / s(t) \big] & = \frac{\vb(\vx_t,t,\vx^r_t)}{s(t)} \\
\vb(\vx_t, t, \vx_t^r) &= s(t) \frac{\mathrm{d}}{\mathrm{d}t} \big[ \hat{\vb}(\vx_t, t, \vx_t^r) / s(t) \big] 
\end{align}
This completes the proof.

\section{Unbiasedness of CoFlow}
\label{AppendixC}
In this section, we provide a rigorous mathematical proof that the proposed CoFlow interpolant strictly guarantees an unbiased transformation from the prior noise distribution $p_1$ to the target data distribution $p_0$, provided that the repulsive adjustment term respects the boundary conditions.

Let $\vx_0 \sim p_0$ be the data distribution and $\vx_1 \sim p_1$ be the prior noise distribution. Let $\vz = (\vx_0, \vx_1, \vx_0^r, \vx_1^r)$ denote the joint set of base and reference samples. We define the CoFlow stochastic interpolant path as:$$\hat{\vx}_t = I(t, z) = s(t)\vx_0 + s(t)\sigma(t)\vx_1 - \hat{\vb}_t(\vz)$$where $\hat{\vb}_t(\vz)$ is an abbreviation for the repulsive adjustment term $\hat{\vb}(\vx_t, t, \vx_t^r)$. For the CoFlow adjustment term, we strictly enforce $\hat{\vb}_0(\vz) = 0$ and $\hat{\vb}_1(\vz) = 0$ for all $\vz$.Let $\rho_t(x)$ denote the time-dependent marginal probability density function of $\hat{\vx}_t$.

First, we check the boundary situation when $t=0$ and $t=1$:
\begin{equation}
    \hat{\vx}_0 = 1 \cdot \vx_0 + 0 \cdot \vx_1 - \hat{\vb}_0(\vz) = \vx_0
\end{equation}

Since $\vx_0 \sim p_0$, the marginal density $\rho_0(\vx) = p_0(\vx)$.
\begin{equation}
   \hat{\vx}_1 = 0 \cdot \vx_0 + 1 \cdot \vx_1 - \hat{\vb}_1(\vz) = \vx_1 
\end{equation}

Since $\vx_1 \sim p_1$, the marginal density $\rho_1(\vx) = p_1(\vx)$. 

About the flow from $\rho_1$ to $\rho_0$, the continuty equation always holds:
\begin{equation}
    \frac{\partial \rho_t(\vx)}{\partial t} + \nabla \cdot (\rho_t(\vx) \vv_t(\vx)) = 0
\end{equation}
Therefore, once $\vv_t(x)$ is obtained, we can use it to transfer between the two distributions. To solve $\vv_t(x)$, we use the standard test function approach. For any test function $\phi(\vx) $, we consider the time derivative of its expectation with respect to the marginal density $\rho_t(x)$:
\begin{equation}
   \frac{d}{dt} \mathbb{E}_{\hat{\vx}_t \sim \rho_t} [\phi(\hat{\vx}_t)] = \frac{d}{dt} \int \phi(\vx) \rho_t(\vx) d\vx = \int\phi(\vx) \frac{\partial \rho_t(\vx)}{\partial t} d\vx
\end{equation}
Alternatively, by taking the derivative inside the expectation over the path generation process and applying the chain rule:
\begin{align}
\frac{d}{dt} \mathbb{E}_{\vx_t} [\phi(\hat{\vx}_t)] &= \mathbb{E}_{\vx_t} \left[ \nabla \phi(\hat{\vx}_t) \cdot \frac{d\hat{\vx}_t}{dt} \right]    \\
 & = \mathbb{E}_{\hat{\vx}_t \sim \rho_t} \left[ \mathbb{E}_\vz \left[ \nabla \phi(\hat{\vx}_t) \cdot \frac{d\hat{\vx}_t}{dt} \mathrel{\bigg|} \hat{\vx}_t \right] \right] \\
 & = \mathbb{E}_{\hat{\vx}_t \sim \rho_t} \left[ \nabla \phi(\hat{\vx}_t) \cdot \mathbb{E}_\vz \left[ \frac{d\hat{\vx}_t}{dt} \mathrel{\bigg|} \hat{\vx}_t \right] \right] \\
 & = \int \rho_t(\vx) \nabla \phi(\vx) \cdot \vv_t(\vx) d\vx \\
 & = - \int\phi(\vx) \nabla \cdot (\rho_t(\vx) \vv_t(\vx)) d\vx
\end{align} 

Here, $\vz$ means all possible $(\vx_0, \vx_1)$ pairings that pass given $\vx_t$, and $\vv_t(\vx) = \mathbb{E}_\vz \left[ {d\hat{\vx}_t}/{dt} \mathrel{\big|} \hat{\vx}_t = \vx \right]$.
This means:
\begin{equation}
    \int\phi(\vx) \frac{\partial \rho_t(\vx)}{\partial t} d\vx = - \int\phi(\vx) \nabla \cdot (\rho_t(\vx) \vv_t(\vx)) d\vx
\end{equation} 
This perfectly matches the continuity equation, which means $\vv_t(\vx)$ defined as $\mathbb{E}_\vz \left[ {d\hat{\vx}_t}/{dt} \mathrel{\big|} \hat{\vx}_t = \vx \right]$ is the velocity field that can achieve the distribution transformation.

According to Theorem 7 in Stochastic Interpolant~\citep{albergo2023stochastic}, to learn this marginal velocity field,  a neural network $\vu_\theta(\hat{\vx}_t, t)$ can be trained to minimize the mean squared error against the conditional velocity target:$$\mathcal{L}_{CoFlow}(\theta) = \mathbb{E}_{t, \vz} \left[ || \vu_\theta(\hat{\vx}_t, t) - \frac{d\hat{\vx}_t}{dt} ||^2 \right]$$

Therefore, we prove that the training of CoFlow is the same as that in standard FM.

\section{Training-free Equivalence of DeltaFM}
\label{AppendixD}
\begin{table}[t]
  \centering
  \vspace{-1em}
  
    \renewcommand{\arraystretch}{1.2}
  \caption{C omparison with DeltaFM. Both models are trained with REPA involved and NEF=50 for inference. ``VC'' denotes the inference-time Velocity Centering operation.}
  \label{tab:deltafm}
  
  \begin{tabular}{llccc}
    \toprule
    \textbf{Model} & \textbf{Method} & \textbf{FID} $\downarrow$ & \textbf{sFID} $\downarrow$ & \textbf{IS} $\uparrow$ \\
    \midrule
    \multirow{3}{*}{SiT-B/2} 
     &  \textcolor{gray}{FM}      &  \textcolor{gray}{27.64} &  \textcolor{gray}{11.76} &  \textcolor{gray}{61.07} \\
    & DeltaFM & 20.40 & 5.48 & 70.41 \\
   
    & {FM + VC} &{19.97} & {5.45} & {71.44} \\
    \midrule
    \multirow{3}{*}{SiT-XL/2} 
     & \textcolor{gray}{FM}      &  \textcolor{gray}{11.40} &  \textcolor{gray}{8.69} &  \textcolor{gray}{114.69} \\
    & DeltaFM & 7.29 & 4.93 & 129.89 \\

    & {FM + VC} & {7.38} & {4.84} & {130.25} \\
    \bottomrule
  \end{tabular}
  \vspace{-1em}
\end{table}
In this section, we provide the detailed derivation for the training-free formulation of DeltaFM and discuss the empirical results presented in Table (\ref{tab:deltafm}). We adapt the derivations in the DeltaFM appendix~\citep{stoica2025contrastive} to our notation. The objective of DeltaFM is defined as follows:

\begin{equation*}
\begin{split}
    \mathcal{L}^{(\Delta FM)}(\theta) = \mathbb{E}\left[
    \begin{aligned}
        & ||\vv_\theta(\vx_t, t) - (\dot{\alpha}_t \hat{\vx} + \dot{\sigma}_t \rvepsilon) ||^2 \\
        & -\lambda ||\vv_\theta(\vx_t, t) - (\dot{\alpha}_t \tilde{\vx} + \dot{\sigma}_t \tilde{\rvepsilon}) ||^2
    \end{aligned}
    \right]
\end{split}
\end{equation*}
Let $(\hat{\vx}, \rvepsilon)$ denote the positive data-noise pair used to construct the model's input, $\vx_t = \alpha_t \hat{\vx} + \sigma_t \rvepsilon$. Similarly, let $(\tilde{\vx}, \tilde{\rvepsilon})$ represent another data-noise pair independently sampled from the current batch, serving as the negative sample. The interpolant coefficients $\alpha_t$ and $\sigma_t$ correspond to $s(t)$ and $s(t)\sigma(t)$ in our CoFlow framework, respectively.

For brevity, write $\vv(\theta)=\vv_\theta(\vx_t,t)$.
Expanding the squared norms and completing the square gives
\begin{align}
\begin{split}
    \mathcal{L}^{(\Delta\mathrm{FM})}(\theta)
    ={}&\mathbb{E}\Big[
        (1-\lambda)\|\vv(\theta)\|^2 \\
       &\quad-2\vv(\theta)^\top
        \Big(
        \dot{\alpha}_t\hat{\vx}
        +\dot{\sigma}_t\rvepsilon
        -\lambda(
        \dot{\alpha}_t\tilde{\vx}
        +\dot{\sigma}_t\tilde{\rvepsilon})
        \Big)
        \Big]+C_1,
\end{split}\\
\begin{split}
    ={}&\mathbb{E}\left[
        \left\|
        \sqrt{1-\lambda}\,\vv(\theta)
        -
        \frac{
        \dot{\alpha}_t\hat{\vx}
        +\dot{\sigma}_t\rvepsilon
        -\lambda(
        \dot{\alpha}_t\tilde{\vx}
        +\dot{\sigma}_t\tilde{\rvepsilon})
        }{\sqrt{1-\lambda}}
        \right\|^2
        \right]+C_2,
\end{split}
\end{align}
where $C_1$ and $C_2$ do not depend on $\theta$ and therefore
do not affect the minimizer.

For a fixed input $(\vx,t)$, the conditional objective is
strictly convex in the predicted velocity because
$1-\lambda>0$.
Setting its gradient to zero yields
\begin{equation}
\begin{aligned}
    (1-\lambda)\vv_t^{\Delta\mathrm{FM},*}(\vx)
    ={}&
    \mathbb{E}\left[
        \dot{\alpha}_t\hat{\vx}
        +\dot{\sigma}_t\rvepsilon
        \mid \vx_t=\vx,t
    \right] \\
    &-\lambda\mathbb{E}\left[
        \dot{\alpha}_t\tilde{\vx}
        +\dot{\sigma}_t\tilde{\rvepsilon}
        \mid \vx_t=\vx,t
    \right].
\end{aligned}
\end{equation}

The first conditional expectation is the optimal standard
FM velocity field:
\begin{equation}
    \vv_t^{\mathrm{FM},*}(\vx)
    =
    \mathbb{E}\left[
        \dot{\alpha}_t\hat{\vx}
        +\dot{\sigma}_t\rvepsilon
        \mid \vx_t=\vx,t
    \right].
\end{equation}
Because the negative pair is independent of the positive
pair and the sampled time, the second conditional
expectation simplifies to
\begin{equation}
\begin{aligned}
    &\mathbb{E}\left[
        \dot{\alpha}_t\tilde{\vx}
        +\dot{\sigma}_t\tilde{\rvepsilon}
        \mid \vx_t=\vx,t
    \right] \\
    &\qquad=
    \mathbb{E}\left[
        \dot{\alpha}_t\tilde{\vx}
        +\dot{\sigma}_t\tilde{\rvepsilon}
        \mid t
    \right]
    =\vm_t.
\end{aligned}
\end{equation}
We therefore obtain the following relation between the
optimal velocity fields:
\begin{equation}
    \vv_t^{\Delta\mathrm{FM},*}(\vx)
    =
    \frac{
        \vv_t^{\mathrm{FM},*}(\vx)-\lambda\vm_t
    }{1-\lambda}.
    \label{eq:closed_form}
\end{equation}

Since the Gaussian prior has zero mean, $\vm_t$ simplifies to
\begin{align}
    \vm_t
    &=
    \dot{\alpha}_t
    \mathbb{E}_{\tilde{\vx}\sim p_0}[\tilde{\vx}]
    +
    \dot{\sigma}_t
    \mathbb{E}_{\tilde{\rvepsilon}\sim\mathcal{N}(0,\mI)}
    [\tilde{\rvepsilon}] \\
    &=
    \dot{\alpha}_t\boldsymbol{\mu}_{\mathrm{data}},
\end{align}
where $\boldsymbol{\mu}_{\mathrm{data}}$ is the mean of
the data distribution in VAE latent space. Eq.~(\ref{eq:closed_form}) shows that the optimal DeltaFM
velocity field is an affine transformation
of the optimal standard FM field. Applying this transformation to a learned standard FM velocity field gives an approximation to the optimal DeltaFM field: 
\begin{equation}
    \vu_\theta^{\mathrm{\Delta FM}}(\vx_t)
    =
    \frac{
        \vu_\theta^{\mathrm{FM}}(\vx_t)
        -\lambda\vm_t
    }{1-\lambda},
    \label{eq:vc}
\end{equation}
where $\vu_\theta^{\Delta\mathrm{FM}}$ denotes the transformed
prediction and $\vm_t$ is estimated from the training data.
We call this operation velocity centering (VC), which
requires no additional model training. To verify its effectiveness, we conducted  the experiments in Table~\ref{tab:deltafm}. Sepcifically, we estimated
the data mean by averaging the SD-VAE latents across the
entire training set.

To evaluate VC, we estimated the data mean by averaging
the SD-VAE latents across the entire training set and
applied the transformation to trained FM models during
inference. As shown in Table~\ref{tab:deltafm}, FM+VC achieves
performance comparable to models explicitly trained
with the DeltaFM objective, supporting the effectiveness
of this training-free formulation.

\section{Lipschitz Constants and Euler Discretization Error}
\label{app:euler_error}

Consider the ODE $\dot{\vx}(s)=\vv(\vx(s),s)$. Assume that, in a neighborhood of the trajectory over $[t,t+h]$, the velocity is bounded by $M$ and is Lipschitz continuous in space and time with constants $L$ and $B$, respectively. Starting from the definition of the single-step Euler error, we obtain
\begin{equation}
\begin{aligned}
\mathcal{E}
&= \|\vx(t+h)-\vx(t)-h\vv(\vx(t),t)\| \\
&= \left\|
    \int_t^{t+h}
    \bigl[\vv(\vx(s),s)-\vv(\vx(t),t)\bigr]
    \,\mathrm{d}s
    \right\| \\
&\le \int_t^{t+h}
    \Bigl[
    \|\vv(\vx(s),s)-\vv(\vx(t),s)\|
    + \|\vv(\vx(t),s)-\vv(\vx(t),t)\|
    \Bigr]\,\mathrm{d}s \\
&\le \int_t^{t+h}
    \bigl[L\|\vx(s)-\vx(t)\|+B(s-t)\bigr]
    \,\mathrm{d}s \\
&\le \int_t^{t+h}
    \left[
    L\int_t^s \|\vv(\vx(r),r)\|\,\mathrm{d}r
    +B(s-t)
    \right]\,\mathrm{d}s \\
&\le \int_t^{t+h}(LM+B)(s-t)\,\mathrm{d}s \\
&= \frac{1}{2}(LM+B)h^2.
\end{aligned}
\end{equation}
For fixed $M$ and $B$, reducing the spatial Lipschitz constant $L$ tightens this error bound.

\end{document}